\documentclass[10pt,twocolumn,letterpaper]{article}

\usepackage{iccv}
\usepackage{times}
\usepackage{epsfig}
\usepackage{graphicx}
\usepackage{amsmath}
\usepackage{amssymb}
\usepackage{booktabs}
\usepackage{makecell}

\usepackage[pagebackref=true,breaklinks=true,letterpaper=true,colorlinks,bookmarks=false]{hyperref}

\iccvfinalcopy 

\def\iccvPaperID{4} 

\ificcvfinal\fi
\begin{document}

\title{Distantly Supervised Road Segmentation}

\author{Satoshi Tsutsui\thanks{The work was done while the author was an intern at Preferred Networks, Inc.}\\
Indiana University\\
{\tt\small stsutsui@indiana.edu}
\and
Tommi Kerola\\
Preferred Networks, Inc.\\
{\tt\small tommi@preferred.jp}
\and
Shunta Saito\\
Preferred Networks, Inc.\\
{\tt\small shunta@preferred.jp}
}

\maketitle
\thispagestyle{empty}

\begin{abstract}
We present an approach for road segmentation that only requires image-level annotations at training time. We leverage distant supervision, which allows us to train our model using images that are different from the target domain. Using large publicly available image databases as distant supervisors, we develop a simple method to automatically generate weak pixel-wise road masks. These are used to iteratively train a fully convolutional neural network, which produces our final segmentation model. We evaluate our method on the Cityscapes dataset, where we compare it with a fully supervised approach. Further, we discuss the trade-off between annotation cost and performance. Overall, our distantly supervised approach achieves $93.8$\% of the performance of the fully supervised approach, while using orders of magnitude less annotation work.
\end{abstract}


\section{Introduction}


Classifying each pixel as corresponding to a road or not is an essential task for practical autonomous driving systems. While basic research on fine-grained segmentation for autonomous driving has been explored previously~\cite{cordts2016cityscapes}, a practical driving system does not necessarily require segmentation (i.e., pixel-wise classification) for most static or moving traffic objects. For objects such as vehicles, pedestrians, and traffic signs, localizing them with bounding boxes is sufficient for achieving satisfactory information for the autonomous driving task. The pixel-wise classification is, however, necessary for the road itself, which is the region where a car is able to drive safely. With this motivation in mind, this paper focuses on segmenting road in an image taken from a \emph{car centric} image. That is, an image taken from a monocular front-facing camera placed on the ego-car. We note that a car centric road image is fundamentally different from a \emph{general} road image, which can be taken from various perspectives, including aerial views. A car centric image, on the other hand, always has the same perspective view, where the car body is clearly visible as a constant part of each image. 

\begin{figure}[t!]
  \centering
  \includegraphics[width=80mm]{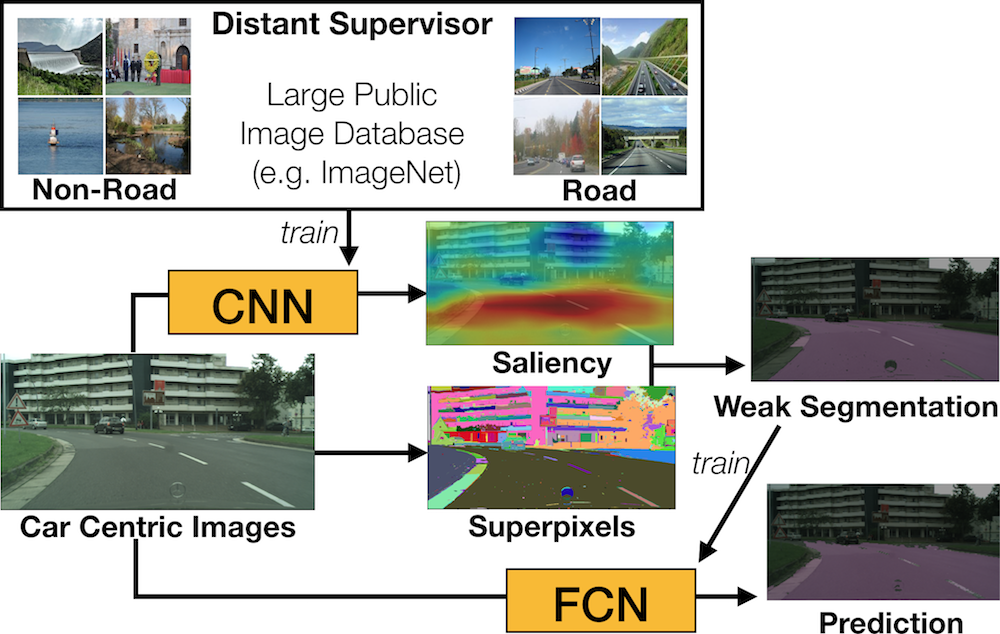}
 	\caption{Overview of our method. From a car centric image, we extract superpixels and saliency that is trained from general images different from car centric images. These are then combined to generate an initial weak segmentation mask. This mask is then used for training an FCN, which results in an accurate road segmentation result. Our method operates without any pixel-wise annotations and only requires a set of image-level labels at training time.}
	\label{fig:overview}
\end{figure}

State-of-the-art segmentation algorithms use fully convolutional neural networks (FCNs), assuming that an abundance of annotated images is available for training~\cite{BadrinarayananSegnet,zhao2016pyramid,lin2016refinenet}. While FCNs can benefit from the progress of modern GPUs or optimized chips, the cost for obtaining annotations is still a bottleneck, requiring much manual work. The cost for segmentation annotation is actually one of the highest compared to other vision-related tasks. For example, the cost to obtain a segmentation mask annotation for an object in an image is reported to be about $80$ seconds, whereas an image label annotation only requires one second~\cite{Bearman2016}. It is also reported that pixel-level road image annotation needs $1.5$ hours per image, including quality control~\cite{cordts2016cityscapes}. This is our motivation to develop a method that is able to train FCNs without using pixel-wise annotations. In this way, the annotation cost is greatly reduced by enabling training with only image-label labels, which is called weakly supervised segmentation. 

Recent work on weakly supervised segmentation employs a convolutional neural network (CNN) trained with image-level labels for the task of classification~\cite{shimoda2016distinct,kolesnikov2016seed,Durand2017,saleh2016built,shimoda2016distinct}, and utilizes a saliency map that highlights the pixels contributing to the classification results~\cite{Zhou2016Discriminative}. However, this approach cannot directly be applied to road segmentation using car centric images, as in all images, the road is visible, which would make it impossible to train a classifier, since we have no negative samples. Even if we collect non-road images and train a binary classifier, the saliency map would highlight the non-road objects that always appear in the car centric image (e.g, the car body). An idea for circumventing this is to learn saliency making use of existing image databases different from the car centric domain.

This is called learning with \emph{distant supervision}~\cite{mintz2009distant}. Distant supervision was originally proposed for reducing the cost of annotating sentences in natural language processing. This technique automatically annotates sentences, making use of existing databases, such as Wikipedia, and use these possibly noisy or weak annotations for training machine learning algorithms.

In this work, we propose to make use of existing image databases different from the car centric domain to train FCNs for road segmentation, and empirically show its performance and challenges. We collect road and non-road images different from the car centric images, by selecting labels from the ImageNet~\cite{Russakovsky2015ImageNet} and Places~\cite{zhou2017places} datasets. Thus, our practical annotation cost boils down to selecting labels from these databases. The selected images are used for training CNNs that can be used for obtaining saliency maps for target road images. These saliency maps are used to generate (noisy and weak) pixel-wise annotations, which are used for training FCNs for road segmentation. It should be noted that our approach is not dependent on the specific architecture of FCNs, nor does it require any modification to it. 

Our experiments on the Cityscapes dataset~\cite{cordts2016cityscapes} show that our approach can achieve $93.8\%$ of the performance of a fully-supervised approach, while requiring orders of magnitude less annotation work. To the best of our knowledge, this is the first paper that reports the performance of a weakly supervised approach to train FCNs for road segmentation.

In summary, we make the following contributions:
\begin{itemize}
	\setlength{\parskip}{0cm} 
	\setlength{\itemsep}{0cm} 
	\item We propose an approach for distantly supervised road segmentation, which does not require any annotation of car centric images, making use of existing image-level label databases.
	\item Using the Cityscapes dataset~\cite{cordts2016cityscapes}, we compare our approach with fully supervised segmentation, provide quantitative and qualitative evaluations, and discuss the trade-off between annotation cost and performance.
\end{itemize}

\section{Related Work}
\paragraph{Fully supervised semantic segmentation}
Semantic segmentation of general objects has been a major part of the computer vision community ever since the PASCAL VOC challenge in 2007~\cite{Everingham2014ThePV}. While various methods have been proposed for this task, a significant breakthrough came in 2014, when end-to-end learning with FCNs was shown to be possible for this task~\cite{long2015fully}. Their general framework for end-to-end semantic segmentation was formulated as an encoder network followed by a decoder network. SegNet~\cite{BadrinarayananSegnet} improved upon this work by introducing an upsampling method that reflects the pooling indices used in the encoder network. This resulted in higher quality output feature maps. Also, since no parameters are used in the upsampling step, this network is more light-weight than the previous methods at the time. U-Net~\cite{ronneberger2015u} is another recent method that uses a ladder-like structure for concatenating feature maps from the encoder and decoder, together with skip connections that allow the decoder to keep details that were previously lost when being pooled in the decoder.
Later, the fully convolutional DenseNet~\cite{jegou2016one} was proposed, which extended U-Net by using the parameter-efficient DenseNet~\cite{huang2016densely} as its base encoder, which resulted in a lightweight model that still achieves high quality segmentation results.
PSPNet~\cite{zhao2016pyramid} steered away from the need of downsampling in the encoder network, and instead used dilated convolutions to keep the feature map at a high resolution. Global context was then aggregated by leveraging spatial pyramid pooling.
Contrary to this, RefineNet~\cite{lin2016refinenet} proposed to use long-range residual connections in order to exploit all available information throughout the downsampling process. This enabled deeper layers to capture fine-grained semantic information by using the convolution input directly from the earlier layers. 


\paragraph{Weakly supervised semantic segmentation}
Distantly supervised segmentation falls into a category of weakly supervised segmentation. Previous work utilizes various weak supervisors such as points \cite{Bearman2016}, scribbles \cite{Lin2016}, bounding boxes \cite{Khoreva2017}, or just image-level labels \cite{shimoda2016distinct,kolesnikov2016seed,Durand2017,saleh2016built,shimoda2016distinct}. In this work, we are interested in the last one: learning segmentation models from just image-level labels. Here we review FCN-based approaches as they are the backbone of the state-of-the-art results. 
  
Early studies view the problem as a multiple instance learning~\cite{pinheiro2015image,pathakICCV15ccnn}. In this problem, a label is assigned to a set of features, and the pixel level predictions are aggregated into per image prediction losses~\cite{andrews2003multiple}. However, it fails to consider object localization information that is implicitly learned by an FCN, which is done in the following research as well as in our approach~\cite{Zhou2016Discriminative}. Saleh~et~al.~\cite{saleh2016built} use an object saliency map (or objectness) learned by classification CNNs. Shimoda and Yanai~\cite{shimoda2016distinct} proposed a method for extracting a class-specific saliency map using gradient-based optimization, and utilized it for segmentation. Kolesnikov and Lampert~\cite{kolesnikov2016seed} proposed several loss functions for weakly supervised segmentation, taking advantage of localization cues learned by FCNs. Durand~et~al.~\cite{Durand2017} proposed an FCN architecture for learning better localization information. 

A related approach to ours is webly supervised semantic segmentation~\cite{Jin2017WeblySS}, which can be regarded as a specific type of distantly supervised segmentation. It collects three sets of images: images of objects with white background, images of common background scenes, and images of objects with common background scenes. These three sets are effectively used to train FCNs. It is effective for segmenting foreground objects (e.g. car) but unlike our approach, it is not applicable for background objects like road, because it is not practical to collect road images with white background. 




\paragraph{Road segmentation}
Road segmentation is often called free space estimation in autonomous driving. Free space is defined as space where a vehicle can drive safely without collision. Free space estimation is often solved by a geometric modeling approach, using more information such as stereo or consistency between frames~\cite{badino2007free,wedel2009b}. The consistency is also used in traditional monocular vision approaches~\cite{alvarez20103d}. Other traditional approaches from  monocular vision employ classifiers based on manually defined features~\cite{alvarez2011road,hanisch2017free}. These features are learned by CNNs in modern approaches. The early work with CNNs utilizes a generic segmentation dataset for the road segmentation problem~\cite{alvarez2012road}. The CNN at that time was patch-based, and was not as sophisticated as the state-of-the-art FCNs. Oliveira~et~al.~\cite{oliveira2016efficient} investigated the performance of modern CNNs for road segmentation, and demonstrated its efficiency. While these approaches require pixel-wise annotations, we focus on the problem of training a road segmentation CNN only with image label annotations.


\section{Method}\label{sec:method}
This section describes our approach for distantly supervised road segmentation using only image-level labels. The key point is that we make use of publicly available large image databases, 
do not need to annotate the car centric images, and only need to select some labels that correspond to road and non-road. After acquiring road and non-road images as distant supervisors, we use these for obtaining initial weak segmentation masks. For the initial segmentation, we employ a very simple approach in order to explore the performance of distant supervision for road segmentation. We note that while previous work~\cite{Durand2017,kolesnikov2016seed} provide more sophisticated methods to train FCNs from image-level labels, we intentionally use a much simpler method in this work, as it is enough for illustrating our framework's proof of concept.

Our initial segmentation masks are obtained by combining a saliency map and superpixels. The saliency map is acquired from a road classification CNN. The superpixels are needed because the saliency map is not sufficiently fine-grained for generating a sharp segmentation mask. Finally, the weak road segmentation result is used as weak labels for training an FCN for generating higher quality road segmentation masks. Each of the steps are explained in detail in the following paragraphs. 

\paragraph{Collect road and non-road images}
We collect road and non-road images for training a road image classification CNN. Non-road images are needed because all the images taken by a camera mounted in the ego-vehicle will have visible road. But the ego-vehicle images cannot be used for representing the road class, since they are quite homogeneous, non-road objects such as the car body will be salient if we train a classifier on them (as our experimental results will show, this holds true). For this reason, we collect a set of non-homogeneous road images as well. Instead of collecting and annotating images from scratch, we propose to use a distant supervision approach, and take advantage of large publicly available image databases~\cite{Russakovsky2015ImageNet,zhou2017places}. The specific data collection process is described in Section~\ref{sec:image_collection_saliency}. 

\paragraph{Train a saliency map generator from image labels}
After we collect the road and non-road images, we train a classification CNN so that we can obtain a road saliency map. For this task, we choose to leverage a CNN with global average pooling (GAP)~\cite{Zhou2016Discriminative} as a simple architecture for obtaining a saliency map.
Let $f_k(x,y)$ denote the activation of channel $k$ in the spatial position $(x,y)$ in the last feature map of the CNN. Then, the score $S_c$ for class $c$ is defined as

\begin{align}
S_c = \frac1N \sum_{k} w_k^c \underbrace{\sum_{x,y}  f_k(x, y)}_{F^k} = \frac1N \sum_{x,y} \underbrace{\sum_k w_k^c f_k(x, y)}_{M_c(x,y)}~,
\end{align}
\noindent where $N$ is the number of spatial positions, $F^k$ is the result of global average pooling, and the class-specific weights $w_k^c$ are learned during training. The term $M_c(x,y)$ can then be interpreted as the saliency for class $c$ at the spatial location $(x,y)$~\cite{Zhou2016Discriminative}.

\begin{figure}[t!]
  \centering
  \includegraphics[width=60mm]{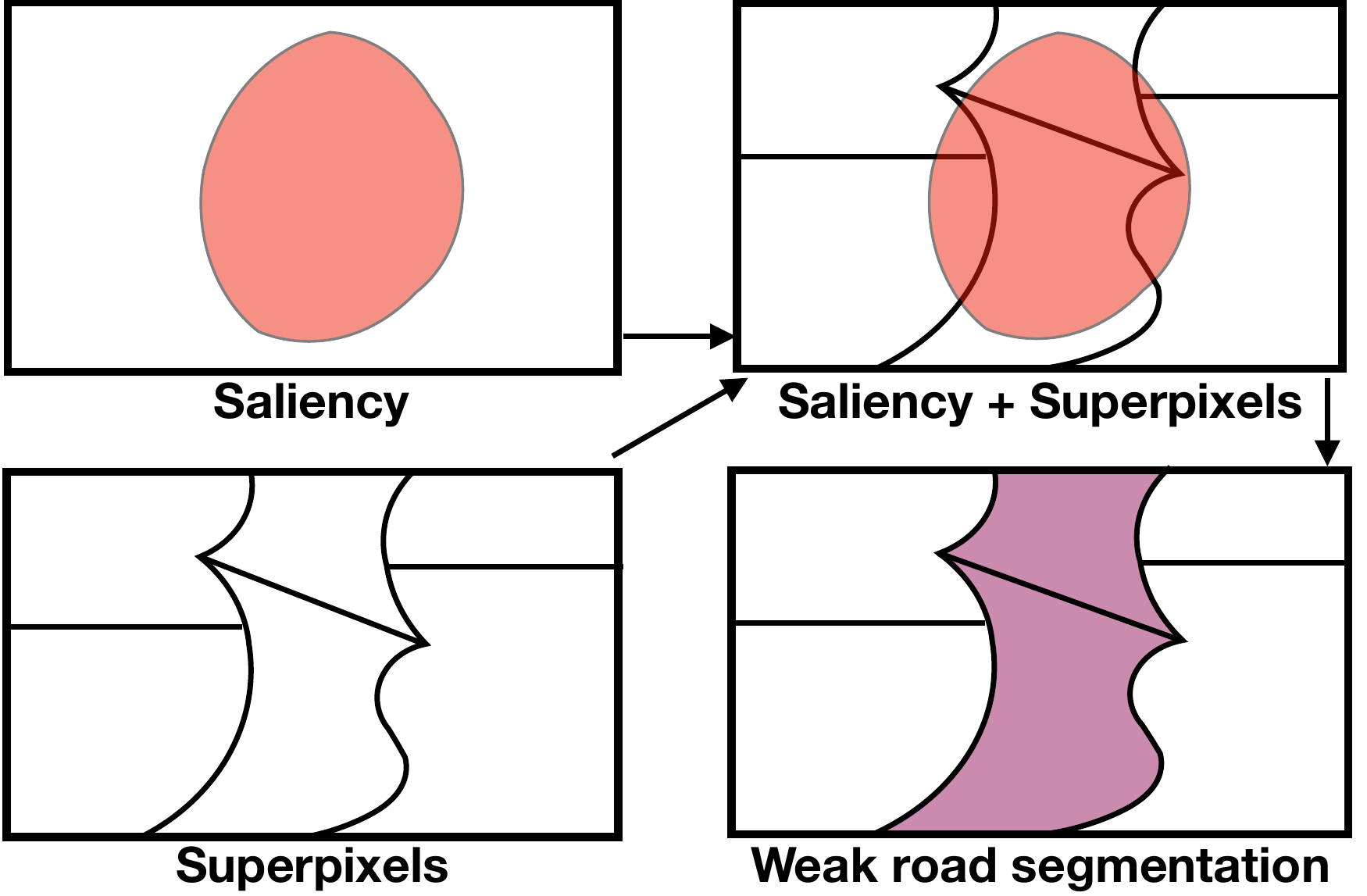}
 	\caption{A weak road segmentation mask is obtained by marking superpixels with sufficient saliency overlap as belonging to the road.}
	\label{fig:method}
\end{figure}

\begin{figure}[t!]
  \centering
  \includegraphics[width=85mm]{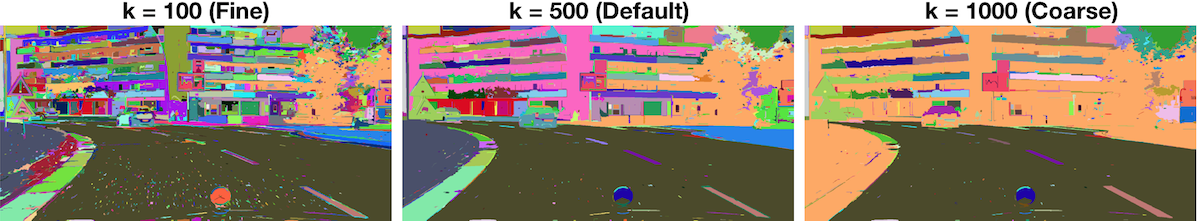}
 	\caption{Graph-based superpixels~\cite{felzenszwalb2004efficient}. The threshold parameter $k$ decides how coarsely the image will be segmented by the superpixels.}
	\label{fig:superpixel-samples}
\end{figure}

\paragraph{Obtain weak labels from the saliency and superpixels}
In addition to the saliency map, we use superpixels to supplement the road segmentation results. This process is illustrated in Figure~\ref{fig:method}.
We use a graph-based algorithm for generating the superpixels~\cite{felzenszwalb2004efficient}. The superpixel generation algorithm has a threshold parameter $k$, which decides to which extent the image will be oversegmented (small $k$) or undersegmented (large $k$). Examples of superpixels can be seen in Figure~\ref{fig:superpixel-samples}.
We adapt superpixels based on the assumptions that the saliency map is blurred and not sharp enough to be regarded as an accurate segmentation, and that a large superpixel will cover the road because it tends to have similar appearance (e.g, color and texture) within an image. These assumptions lead us to the following method for combining the saliency map and superpixels in order to obtain better road segmentation results. Let $P_\tau = \{ (x,y) : M(x,y) > \tau \}$ denote the salient area given a saliency threshold $\tau$. Given a set of superpixels $\mathcal{S}$, the weak label $y_\text{weak}^{(i)}$ at location $i$ is defined as

\begin{align}
y_\text{weak}^{(i)} =&
\begin{cases}
    \text{road} & |s_i \cap P_\tau| / |P_\tau| > \theta \\
    \text{other} & \text{otherwise}
\end{cases}~,~\forall s_i \in \mathcal{S}~,
\end{align}
\noindent where $\theta$ is an overlap threshold.
That is, for each superpixel, if the overlap with the salient area is greater than $\theta$, it is regarded as corresponding to a road area. As our experiments will show, this simple strategy of combining saliency and superpixels gives a larger performance gain compared to just using the saliency map (see Section~\ref{sec:weak-road-infer}).

\paragraph{Train an FCN for road segmentation using weak labels} 
We train an FCN~\cite{BadrinarayananSegnet} from the weak labels, where our assumption is that the FCN is able to eliminate the noise of the segmentation results given by the weak labels, and is able to result in a better segmentation mask than the original training data. As our experimental results will show, our assumption holds true (see Section~\ref{sec:train-cnn}).

\section{Experiments}
 
\paragraph{Dataset.} We conduct experiments on Cityscapes~\cite{cordts2016cityscapes}, which is an established dataset proposed for the task of fine-grained general semantic segmentation for autonomous driving. We only use the road class for the evaluation and report the mean intersection over union (mIOU) in a pixel-wise manner, ignoring the void regions defined in the ground truth. We train on $2,975$ training images defined in Cityscapes, and test the performance on $500$ validation images, since ground truth for the test set is not available.

\subsection{Saliency Map by Global Average Pooling}\label{sec:image_collection_saliency}
This section describes an experiment we conducted in order to see the performance of the saliency map obtained by global average pooling (GAP)~\cite{Zhou2016Discriminative}, as well as how we actually collected training images.

\begin{figure}[t!]
  \centering
  \includegraphics[width=85mm]{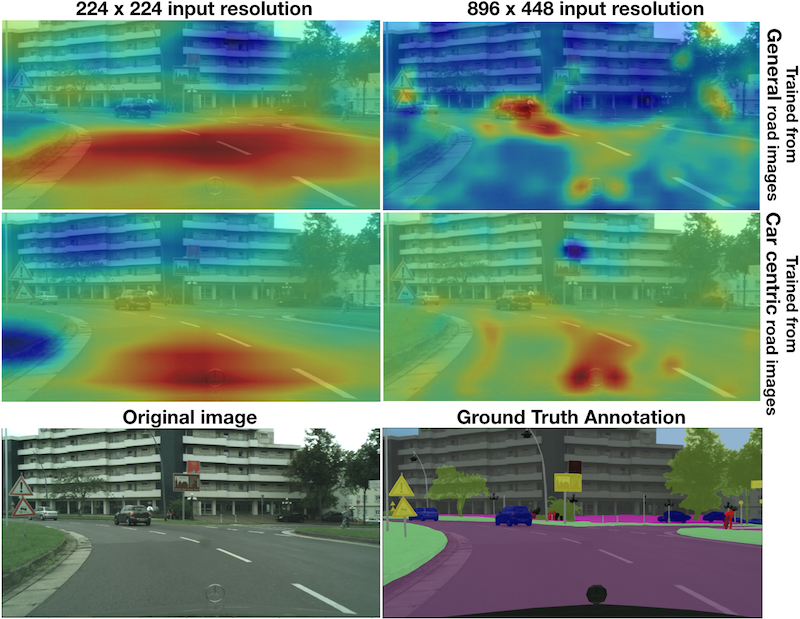}
 	\caption{Examples of saliency estimation difference when using different resolution and/or training images. The best saliency is gotten by training on general images with low resolution. When training on general images with high resolution, cars become salient. When trained from car centric images with low resolution, the car body becomes salient. This tendency is stronger when trained with higher resolution. Note that the bottom black region in the ground truth indicates the car body.}
	\label{fig:saliency}
\end{figure}

\begin{figure}[t!]
  \centering
  \includegraphics[width=85mm]{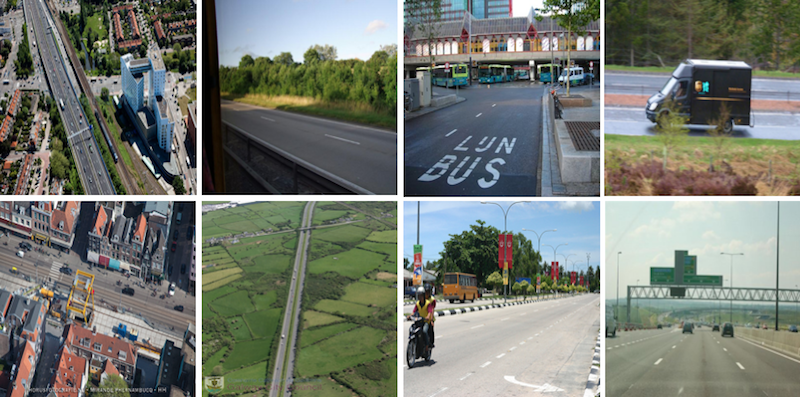}
 	\caption{Examples of road images from ImageNet used for training our saliency detector.}
	\label{fig:road-samples}
\end{figure}

We require an image collection of road and non-road images for training a classification CNN with GAP. Instead of annotating from scratch, we take advantage of two publicly available large image databases: ImageNet~\cite{Russakovsky2015ImageNet} and Places~\cite{zhou2017places}. For road images, we use ImageNet for collecting road images, as its labels are organized in an object centric way. We searched for labels with the keyword \textit{road} or \textit{highway}, which yielded the classes \textit{n02744323: arterial road} and \textit{n02744323: divided highway, dual carriageway}. Random samples are shown in Figure~\ref{fig:road-samples}. 

\begin{figure}[t!]
  \centering
  \includegraphics[width=85mm]{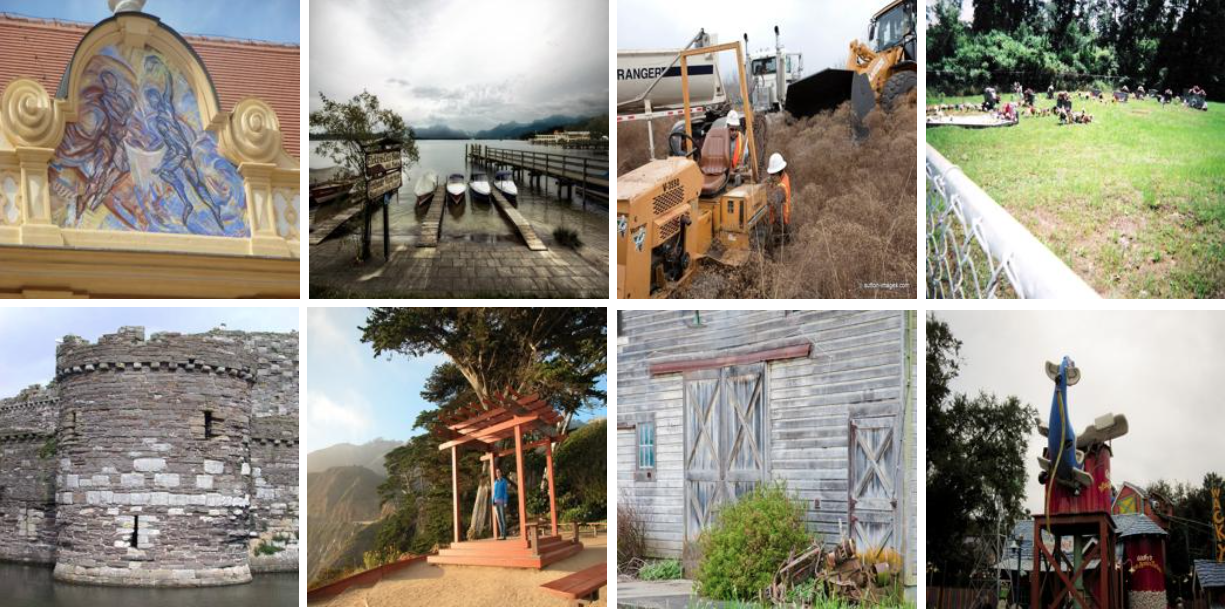}
 	\caption{Examples of non-road images from Places used for training our saliency detector.}
	\label{fig:non-road-samples}
\end{figure}

We do not use the Cityscapes training set as road images, as the saliency FCN would highlight the objects that consistently appear in the Cityscapes images. We prove this empirically as a part of our experiments. For non-road images, we need to collect outdoor scene images without road. We use Places, since unlike ImageNet, it organizes images according to scenes. We first filter out scene labels whose meta class corresponds to indoor scenes, which resulted in a remaining $205$ outdoor labels. We manually examined these labels in order to exclude irrelevant classes (e.g., baseball field) and road classes (e.g., field road), which resulted in $120$ classes. Random samples are shown in Figure~\ref{fig:non-road-samples}.

This is the only practically required manual labeling process in our work. In addition, we do not aim to obtain very accurate images corresponding to a label, as we do not have time to  check all the images manually, so some noisy images are acceptable. For example, the road class contains images that do not look like a typical car centric road, as they are shot from a helicopter (see the left side of Figure~\ref{fig:road-samples}).

We train a VGG-based~\cite{Simonyan2014vgg} architecture pretrained for the ImageNet classification task\cite{Russakovsky2015ImageNet}, replacing fully-connected layers with global average pooling~\cite{Zhou2016Discriminative}. We implemented our models in Python using the Chainer framework~\cite{tokui2015chainer}. While the original method is trained on the resolution of $224 \times 224$ pixels, the images in Cityscapes have a resolution of  $2048 \times 1024$. Therefore, we tried two approaches:
\begin{itemize}
    \item Train using $224 \times 224$ resolution, obtain a $14 \times 14$ saliency map (VGG downsamples the image by a factor of 16).
    \item Train using $896 \times 448$ resolution (higher is not possible due to GPU memory constraints), obtain a $56 \times 28$ saliency map.
\end{itemize}
We then use bilinear interpolation on the saliency map to restore the original Cityscapes resolution.

We evaluate the quality of the saliency map by comparing it with the ground truth road segmentation mask. We set the saliency threshold $\tau = 0.75$, and compute mIOU for the ground truth road in the Cityscapes training set (note that the saliency generator was trained on ImageNet+Places; not Cityscapes). The results are shown in Table~\ref{tbl:saliency}. Perhaps surprisingly, the lower resolution input yields a much better mIOU of $0.405$ compared to $0.092$ for the higher resolution case.  Our manual inspection of some random samples indicated that the higher resolution saliency map tends to highlight objects that often appear on the road, such as cars or traffic signs. A sample image highlighting a car is shown in top-right side of Figure~\ref{fig:saliency}. This makes sense because our approach is data driven, but our findings suggest that we are able to avoid this issue by using lower resolution images. 

In order to prove our argument that we cannot directly use Cityscapes images for road images, we trained a CNN where we replaced road images with Cityscapes training images. As shown in Table~\ref{tbl:saliency}, the mIOU is lower compared to just using road images. We also show examples in the bottom row of Figure~\ref{fig:saliency}. The salient region is actually the car body and the logo of the car company. This tendency is stronger when the input resolution is higher, which is shown in the bottom right side of the figure. This confirms our argument that we also need to collect general road images in addition to the car centric road images.

\begin{table}[tb]
\centering
\begin{tabular}{cccc}
\toprule
    \makecell{Method} &  \makecell{mIOU} \\
\midrule
  Low resolution, Generic road image & 0.405 \\
  High resolution, Generic road image & 0.092 \\
  Low resolution, Car centric road image  & 0.206  \\
  High resolution, Car centric road image & 0.093 \\
\bottomrule
\end{tabular}
\vspace{0.5em}
\caption{Experimental results on Cityscapes when using saliency only.
Car centric images are from Cityscapes. General road images are from ImageNet.
}
\label{tbl:saliency}
\end{table}


\subsection{Integration of Saliency and Superpixels}\label{sec:weak-road-infer}
This section reports the results of generating road segmentation masks by combining saliency map and superpixels. As we discussed in Section~\ref{sec:method}, three types of parameters have to be tuned: the superpixel granularity parameter $k$, the saliency threshold $\tau$, and the overlap threshold $\theta$. For superpixels, we tried $k=100$ (expected finer segmentation), $k=500$ (the default value suggested in the author's implementation), and  $k=1000$ (expected coarser segmentation). 
The results are shown in Table~\ref{tbl:threshold-results}.

Integrating saliency and superpixels is beneficial for making better weak labels. We believe the reason is that the road tends to have homogeneous appearance within an image, resulting in a superpixel covering most of it. Moreover, using a high saliency threshold and a low overlap threshold gave the best results.

\begin{table}[tb]
\centering
\begin{tabular}{cccc}
\toprule
    \makecell{Superpixel\\param. $k$} &  \makecell{Saliency\\thres. $\tau$} &  \makecell{Overlap\\thres. $\theta$} &  mIOU \\
\midrule
  $500$ &  $0.90$ &  $0.01$ &      $\mathbf{0.659}$ \\
  $100$ &  $0.90$ &  $0.01$ &      $0.626$ \\
 $1000$ &  $0.90$ &  $0.01$ &      $0.600$ \\
  $100$ &  $0.75$ &  $0.25$ &      $0.565$ \\
  $500$ &  $0.75$ &  $0.25$ &      $0.505$ \\
  $100$ &  $0.50$ &  $0.50$ &      $0.491$ \\
  $500$ &  $0.50$ &  $0.50$ &      $0.471$ \\
 $1000$ &  $0.50$ &  $0.50$ &      $0.441$ \\
 $1000$ &  $0.75$ &  $0.25$ &      $0.425$ \\
  $100$ &  $0.90$ &  $0.10$ &      $0.414$ \\
  $500$ &  $0.90$ &  $0.10$ &      $0.358$ \\
 $1000$ &  $0.90$ &  $0.10$ &      $0.282$ \\
\bottomrule
\end{tabular}
\vspace{0.5em}
\caption{Experimental results on Cityscapes of tuning our method's threshold parameters.}
\label{tbl:threshold-results}
\end{table}

ƒ
\subsection{Training an FCN from Weak Labels}\label{sec:train-cnn}

\begin{table}[tb]
\centering
\begin{tabular}{ccc}
\toprule
    \makecell{Trained from} & \makecell{mIOU} &\makecell{estim. labeling\\cost (hours)} \\
\midrule
  Weak labels (WL) (iter. 0.) & $0.779$ & $0.6$  \\
  Predictions (iter. 1.)  & $0.790$ & $0.6$   \\
  Predictions (iter. 2.) & $0.798$ & $0.6$   \\
  Predictions (iter. 3.) & $0.800$ & $0.6$   \\
  Predictions (iter. 4.) & $0.799$ & $0.6$   \\
  WL + $60$\% of ground truth  & $0.855$  & $39.7$  \\
  WL + ground truth  & $0.863$  & $65.9$  \\ \hline
  \makecell{Ground truth (baseline)} & $0.853$ & $65.3$  \\
\bottomrule
\end{tabular}
\vspace{0.5em}
\caption{Experimental results on Cityscapes when training an FCN on our generated weak labels. Iter. $k$ is trained from the prediction of iter $k-1$ , except iter. 0, which is trained from weak labels}
\label{tbl:weak-label-results}
\end{table}


This section describes the road segmentation results obtained by training an FCN on our generated weak labels. We use SegNet~\cite{BadrinarayananSegnet} as our choice of FCN, but we note that our method is not dependent on the FCN, so any FCN could be used. We first train SegNet on the segmentation results gotten by saliency and superpixels. We evaluate our results both in quantitative and qualitative manners, regarding the model trained from the ground truth pixel-wise annotations as a baseline. Moreover, we report results of several other experiments that we conducted in order to understand what will improve the performance. The results of the following experiments are summarized in Table \ref{tbl:weak-label-results}.

\paragraph{Quantitative Evaluation}
We evaluate the performance using mIOU on the evaluation images. Training from the weak segmentation resulted in an mIOU of $0.779$, which is $91.3$\% of the fully supervised performance of $0.853$.  Moreover, we note that the mIOU for training images is $0.777$, which is much higher than the mIOU of $0.659$ that was gotten from the initial weak segmentation results. This indicates that training an FCN from weak and noisy labels is effective for removing the outliers and/or noise, and that the FCN is able to learn the common ingredients of the noisy segmentation masks, which should be closer to the ground truth.

\begin{figure}[t!]
  \centering
  \includegraphics[width=85mm]{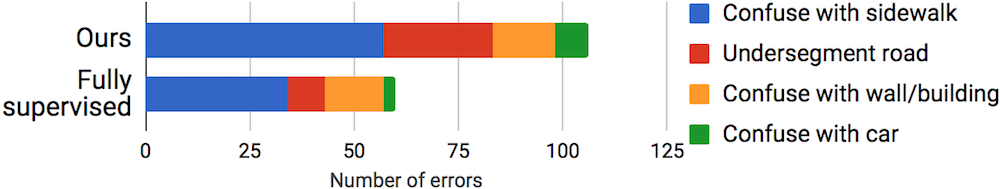}
 	\caption{Error analysis.}
	\label{fig:error-analysis}
\end{figure}

\begin{figure*}[t]
  \centering
  \includegraphics[width=170mm]{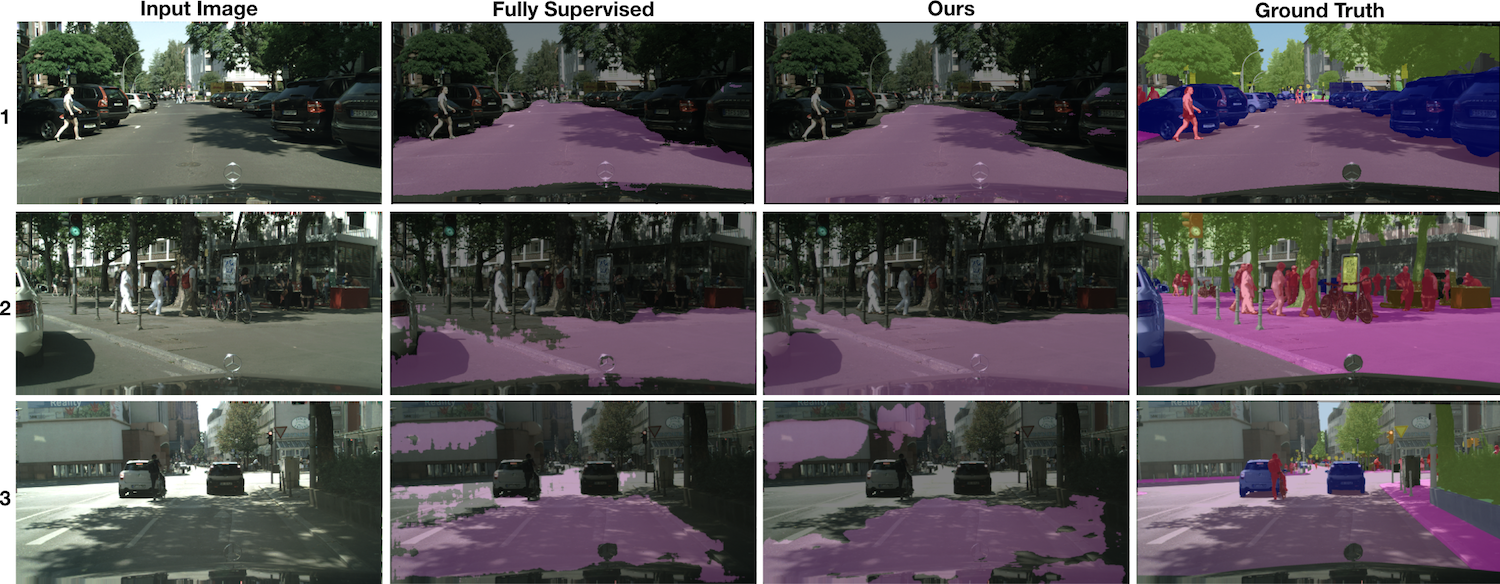}
 	\caption{Sample results, comparing our method with the fully supervised approach and ground truth. Row 1 shows the case where segmentation is relatively successful. In row 2, we selected a sample whose number of false positive pixels are the largest due to incorrectly recognizing sidewalk as road. In row 3, we selected a sample whose number of false negative pixels are the largest. It has large under-segmented area on the road. It also incorrectly recognized building/walls as road.}
	\label{fig:samples}
	\vspace{-0.35cm}
\end{figure*}

\paragraph{Qualitative Evaluation}
We also perform qualitative evaluations by manually inspecting $100$ random samples from the evaluation images, and compared with the ground truth and the segmentation result of the fully supervised model. For each image, we identify the major reason of the performance drop, and summarize into four types. The four types are shown in Figure~\ref{fig:error-analysis}. The evaluation indicates that sidewalks are hard to distinguish from the road for both our approach and the fully supervised approach. This is reasonable, because sidewalks have a very similar appearance to road and require context in order to be distinguished. We also noticed that our approach sometimes yields far more undersegmented results compared to the fully supervised method. This indicates the lower quality of our weak labels, compared to the ground truth. Several examples are shown in Figure~\ref{fig:samples}.

\paragraph{Training on the predicted output}
We confirmed that training on the weak segmentation mask improves the segmentation performance. Therefore, training another FCN using the output of the first FCN as training data might improve the results. Implementing this idea results in an mIOU of $0.790$ on evaluation data, which confirms our hypothesis.
We train yet several more on the output of the previous FCN, and achieve $0.80$ mIOU.
With this training procedure, we are able to match $93.8$\% of the performance of the fully supervised model.

\paragraph{Fine-tune with pixel-wise ground truth annotations}
Assuming that it is difficult to reach the fully supervised performance only with the image label annotations, a practical scenario is to use these for pretraining the FCN, and fine-tune with the ground truth annotation. It would help us to reduce the annotation cost if this requires a smaller number of ground truth images in order to match the fully supervised model. We gradually decrease the number of ground truth images to $90$\%, $80$\%$, $\ldots$, 10$\% and confirm that an mIOU of $0.855$, which is almost tied with the fully supervised model, is reached when using $60$\% of the ground truth labels for fine-tuning. This result suggests that we could obtain better results than a fully supervised model when starting with our model trained from weak labels, and then fine-tuning using all of the ground truth pixel-wise annotations. Implementing this idea resulted in an mIOU of $0.863$.
This shows that we potentially can use our weak labels for low-cost pretraining.

\subsection{Annotation cost estimation}
The fundamental motivation for our work is to reduce the annotation cost. This section estimates the annotation time for our method and the supervised baseline, and discusses the trade-off between annotation time and performance.
The annotation cost for a segmentation mask is estimated to be $79$ seconds for an object in an image~\cite{Bearman2016}. Approximating road in an image as an object, the annotation time for the $2,975$ training images is $2,975 \times 79 = 235,025~\text{sec.} \approx 65.3~ \text{hours}$. On the other hand, the annotation cost in our proposed method is to select labels from ImageNet and Places. For ImageNet, we just searched by the two keywords and selected two labels. We estimate this takes only a minute per label, as we did not carefully investigate the search results. For Places, we manually checked $205$ outdoor classes to see if it contains road. We only checked at most $10$ images from each class. If we assume that image-level labeling takes one second per class~\cite{Bearman2016}, then our required annotation time is $10$ seconds per label. Based on these estimations, the annotation time to collect non-road images is $60 \times 2 + 10 \times 205 = 2170~\text{sec.} \approx 0.6~ \text{hours}$. From this estimation, we can conclude that we are able to obtain $93.8$\% of the performance using less than $1$\% of the annotation cost of the fully supervised model (see Table~\ref{tbl:weak-label-results}). We do not consider the time for creating ImageNet and Places, as we regard them as existing public resources and because pre-training from these large databases is common practice in the computer vision community. We also note that this way of comparison has been used previously in the literature for weakly supervised segmentation, where the cost for creating ImageNet is ignored~\cite{Bearman2016}. 

\section{Conclusion and Future Work}
We presented an approach for distantly supervised road segmentation using fully convolutional neural networks, based on saliency and superpixels.
Our experimental results on Cityscapes showed that our method was able to achieve $93.8$\% of the performance of a fully supervised approach using only image-level labels, while significantly reducing the annotation cost. 

In the future, we will focus on finding a way to match the performance of the fully supervised model. Possible approaches for this could include improving the quality of the saliency detector by e.g., making it operate at a higher resolution. Other promising directions include trying more sophisticated weakly supervised segmentation, such as incorporating conditional random fields~\cite{kolesnikov2016seed}. Finally, leveraging geometric context for adding road-specific priors could be explored as well.

\vspace{-0.2cm}
\paragraph{Acknowledgments}
We would like to thank the members of Preferred Networks, Inc.,
particularly Richard Calland, Zornitsa Kostadinova, Masaki Saito, and
Daichi Suzuo for insightful comments and discussions.
\vspace{-0.2cm}

{\small
\bibliographystyle{ieee}
\bibliography{references}

\begin{thebibliography}{10}\itemsep=-1pt

\bibitem{alvarez2012road}
J.~Alvarez, T.~Gevers, Y.~LeCun, and A.~Lopez.
\newblock {Road scene segmentation from a single image}.
\newblock In {\em {ECCV}}, 2012.

\bibitem{alvarez20103d}
J.~M. Alvarez, T.~Gevers, and A.~M. Lopez.
\newblock {3d Scene Priors for Road Detection}.
\newblock In {\em CVPR}, 2010.

\bibitem{alvarez2011road}
J.~M.~{\'A}. Alvarez and A.~M. Lopez.
\newblock {Road detection based on illuminant invariance}.
\newblock {\em IEEE Trans. ITS}, 2011.

\bibitem{andrews2003multiple}
S.~Andrews, I.~Tsochantaridis, and T.~Hofmann.
\newblock {Support vector machines for multiple-instance learning}.
\newblock In {\em {NIPS}}, 2003.

\bibitem{badino2007free}
H.~Badino, U.~Franke, and R.~Mester.
\newblock Free space computation using stochastic occupancy grids and dynamic
  programming.
\newblock In {\em ICCV workshop}, 2007.

\bibitem{BadrinarayananSegnet}
V.~Badrinarayanan, A.~Kendall, and R.~Cipolla.
\newblock {SegNet: A Deep Convolutional Encoder-Decoder Architecture for Image
  Segmentation}.
\newblock {\em PAMI}, 2017.

\bibitem{Bearman2016}
A.~Bearman, O.~Russakovsky, V.~Ferrari, and L.~Fei-Fei.
\newblock {What's the point: Semantic segmentation with point supervision}.
\newblock In {\em ECCV}, 2016.

\bibitem{cordts2016cityscapes}
M.~Cordts, M.~Omran, S.~Ramos, T.~Rehfeld, M.~Enzweiler, R.~Benenson,
  U.~Franke, S.~Roth, and B.~Schiele.
\newblock {The cityscapes dataset for semantic urban scene understanding}.
\newblock In {\em {CVPR}}, 2016.

\bibitem{Durand2017}
T.~Durand, T.~Mordan, N.~Thome, and M.~Cord.
\newblock {WILDCAT : Weakly Supervised Learning of Deep ConvNets for Image
  Classification , Pointwise Localization and Segmentation}.
\newblock In {\em CVPR}, 2017.

\bibitem{Everingham2014ThePV}
M.~Everingham, S.~A. Eslami, L.~Van~Gool, C.~K. Williams, J.~Winn, and
  A.~Zisserman.
\newblock {The pascal visual object classes challenge: A retrospective}.
\newblock {\em {IJCV}}, 111(1):98--136, 2015.

\bibitem{felzenszwalb2004efficient}
P.~F. Felzenszwalb and D.~P. Huttenlocher.
\newblock {Efficient graph-based image segmentation}.
\newblock {\em IJCV}, 59(2):167--181, 2004.

\bibitem{hanisch2017free}
S.~H{\"a}nisch, R.~H. Evangelio, H.~H. Tadjine, and M.~P{\"a}tzold.
\newblock {Free-Space Detection with Fish-Eye Cameras}.
\newblock In {\em {IV}}, 2017.

\bibitem{huang2016densely}
G.~Huang, Z.~Liu, L.~van~der Maaten, and K.~Q. Weinberger.
\newblock {Densely connected convolutional networks}.
\newblock In {\em CVPR}, 2017.

\bibitem{jegou2016one}
S.~J{\'e}gou, M.~Drozdzal, D.~Vazquez, A.~Romero, and Y.~Bengio.
\newblock {The one hundred layers tiramisu: Fully convolutional DenseNets for
  semantic segmentation}.
\newblock {\em arXiv:1611.09326}, 2016.

\bibitem{Jin2017WeblySS}
B.~Jin, M.~V.~O. Segovia, and S.~S{\"u}sstrunk.
\newblock {Webly Supervised Semantic Segmentation}.
\newblock In {\em CVPR}, 2017.

\bibitem{Khoreva2017}
A.~Khoreva, R.~Benenson, J.~Hosang, M.~Hein, and B.~Schiele.
\newblock {Simple Does It: Weakly Supervised Instance and Semantic
  Segmentation}.
\newblock In {\em CVPR}, 2017.

\bibitem{kolesnikov2016seed}
A.~Kolesnikov and C.~H. Lampert.
\newblock {Seed, Expand and Constrain: Three Principles for Weakly-Supervised
  Image Segmentation}.
\newblock In {\em {ECCV}}, 2016.

\bibitem{Lin2016}
D.~Lin, J.~Dai, J.~Jia, K.~He, and J.~Sun.
\newblock {ScribbleSup: Scribble-Supervised Convolutional Networks for Semantic
  Segmentation}.
\newblock In {\em CVPR}, 2016.

\bibitem{lin2016refinenet}
G.~Lin, A.~Milan, C.~Shen, and I.~Reid.
\newblock {Refinenet: Multi-path refinement networks with identity mappings for
  high-resolution semantic segmentation}.
\newblock In {\em CVPR}, 2017.

\bibitem{long2015fully}
J.~Long, E.~Shelhamer, and T.~Darrell.
\newblock {Fully convolutional networks for semantic segmentation}.
\newblock In {\em CVPR}, 2015.

\bibitem{mintz2009distant}
M.~Mintz, S.~Bills, R.~Snow, and D.~Jurafsky.
\newblock Distant supervision for relation extraction without labeled data.
\newblock In {\em ACL-IJCNLP}, 2009.

\bibitem{oliveira2016efficient}
G.~L. Oliveira, W.~Burgard, and T.~Brox.
\newblock Efficient deep models for monocular road segmentation.
\newblock In {\em {IROS}}, 2016.

\bibitem{pathakICCV15ccnn}
D.~Pathak, P.~Kr\"ahenb\"uhl, and T.~Darrell.
\newblock {Constrained Convolutional Neural Networks for Weakly Supervised
  Segmentation}.
\newblock In {\em {ICCV}}, 2015.

\bibitem{pinheiro2015image}
P.~O. Pinheiro and R.~Collobert.
\newblock From image-level to pixel-level labeling with convolutional networks.
\newblock In {\em {CVPR}}, 2015.

\bibitem{ronneberger2015u}
O.~Ronneberger, P.~Fischer, and T.~Brox.
\newblock {U-net: Convolutional networks for biomedical image segmentation}.
\newblock In {\em MICCAI}, 2015.

\bibitem{Russakovsky2015ImageNet}
O.~Russakovsky, J.~Deng, H.~Su, J.~Krause, S.~Satheesh, S.~Ma, Z.~Huang,
  A.~Karpathy, A.~Khosla, M.~S. Bernstein, A.~C. Berg, and L.~Fei-Fei.
\newblock {ImageNet Large Scale Visual Recognition Challenge}.
\newblock {\em IJCV}, 115:211--252, 2015.

\bibitem{saleh2016built}
F.~Saleh, M.~S.~A. Akbarian, M.~Salzmann, L.~Petersson, S.~Gould, and J.~M.
  Alvarez.
\newblock {Built-in foreground/background prior for weakly-supervised semantic
  segmentation}.
\newblock In {\em {ECCV}}, 2016.

\bibitem{shimoda2016distinct}
W.~Shimoda and K.~Yanai.
\newblock {Distinct class-specific saliency maps for weakly supervised semantic
  segmentation}.
\newblock In {\em {ECCV}}, 2016.

\bibitem{Simonyan2014vgg}
K.~Simonyan and A.~Zisserman.
\newblock {Very Deep Convolutional Networks for Large-Scale Image Recognition}.
\newblock In {\em ICLR}, 2014.

\bibitem{tokui2015chainer}
S.~Tokui, K.~Oono, S.~Hido, and J.~Clayton.
\newblock {Chainer: a next-generation open source framework for deep learning}.
\newblock In {\em {NIPS workshop}}, 2015.

\bibitem{wedel2009b}
A.~Wedel, H.~Badino, C.~Rabe, H.~Loose, U.~Franke, and D.~Cremers.
\newblock B-spline modeling of road surfaces with an application to free-space
  estimation.
\newblock {\em IEEE Trans. ITS}, 2009.

\bibitem{zhao2016pyramid}
H.~Zhao, J.~Shi, X.~Qi, X.~Wang, and J.~Jia.
\newblock {Pyramid Scene Parsing Network}.
\newblock In {\em CVPR}, 2017.

\bibitem{Zhou2016Discriminative}
B.~Zhou, A.~Khosla, {\`A}.~Lapedriza, A.~Oliva, and A.~Torralba.
\newblock {Learning Deep Features for Discriminative Localization}.
\newblock {\em CVPR}, 2016.

\bibitem{zhou2017places}
B.~Zhou, A.~Lapedriza, A.~Khosla, A.~Oliva, and A.~Torralba.
\newblock {Places: A 10 million Image Database for Scene Recognition}.
\newblock {\em PAMI}, 2017.

\end{thebibliography}
}

\end{document}